\documentclass{article}
\usepackage{ijcai23}

\usepackage{times}
\usepackage{soul}
\usepackage{url}
\usepackage[hidelinks]{hyperref}
\usepackage[utf8]{inputenc}
\usepackage[small]{caption}
\usepackage{graphicx}
\usepackage{amsmath}
\usepackage{amsthm}
\usepackage{booktabs}
\usepackage{algorithm}
\usepackage{algorithmic}
\usepackage[switch]{lineno}

\usepackage{subfiles}
\usepackage[table]{xcolor}
\usepackage[mathscr]{eucal}
\usepackage{bbding} 
\definecolor{aliceblue}{rgb}{0.94, 0.97, 1.0}

\usepackage{amsthm,amsmath,amssymb}
\usepackage{mathrsfs} 

\usepackage{multirow}

\title{TG-VQA: Ternary Game of Video Question Answering}

\author{
    Hao Li$^{1,3}$\footnotemark[1]\and
    Peng Jin$^{1,3}$\footnotemark[1]\and
    Zesen Cheng$^{1,3}$\and
    Songyang Zhang$^{2}$\and 
    Kai Chen$^{2}$ \and
    Zhennan Wang$^{4}$\and \\
    Chang Liu$^{5}$\footnotemark[2]\and
    Jie Chen$^{1,3,4}$\footnotemark[2] 
    \affiliations
    \small{$^1$School of Electronic and Computer Engineering, Peking University Shenzhen Graduate School, Shenzhen, China}\\
    \small{$^2$Shanghai AI Laboratory, Shanghai, China} \quad \small{$^3$AI for Science (AI4S)-Preferred Program, Peking University Shenzhen Graduate School} \\
    \small{$^4$Peng Cheng Laboratory, Shenzhen, China} \quad \small{$^5$Department of Automation and BNRist, Tsinghua University}
    \emails
    \footnotesize{lihao1984@pku.edu.cn} \quad \footnotesize{\{jp21,  cyanlaser\}@stu.pku.edu.cn} \quad \footnotesize{sy.zhangbuaa@gmail.com} \quad \footnotesize{chenkai@pjlab.org.cn} \\ 
    \footnotesize{wangzhennan2017@email.szu.edu.cn \quad liuchang2022@tsinghua.edu.cn \quad
    chenj@pcl.ac.cn}
}

\begin{document}

\maketitle

\begin{abstract}
Video question answering aims at answering a question about the video content by reasoning the alignment semantics within them. However, since relying heavily on human instructions, \textit{i.e.}, annotations or priors, current contrastive learning-based VideoQA methods remains challenging to perform fine-grained visual-linguistic alignments. In this work, we innovatively resort to game theory, which can simulate complicated relationships among multiple players with specific interaction strategies, \textit{e.g.}, video, question, and answer as ternary players, to achieve fine-grained alignment for VideoQA task. 
Specifically, we carefully design a VideoQA-specific interaction strategy to tailor the characteristics of VideoQA, which can mathematically generate the fine-grained visual-linguistic alignment label without label-intensive efforts.
Our TG-VQA outperforms existing state-of-the-art by a large margin (more than $5\%$) on long-term and short-term VideoQA datasets, verifying its effectiveness and generalization ability. 
Thanks to the guidance of game-theoretic interaction, our model impressively convergences well on limited data~(${10}^4 ~videos$), surpassing most of those pre-trained on large-scale data~($10^7~videos$).
\footnote{\noindent \hyperlink{https://github.com/HowardLi1984/TG-VQA/}{Code Available}.  $*$ Equal Contribution. $\dagger$ Corresponding author.}

\end{abstract}
\section{Introduction}

\begin{figure}[t]
    \centering
    \includegraphics[width=0.48\textwidth]{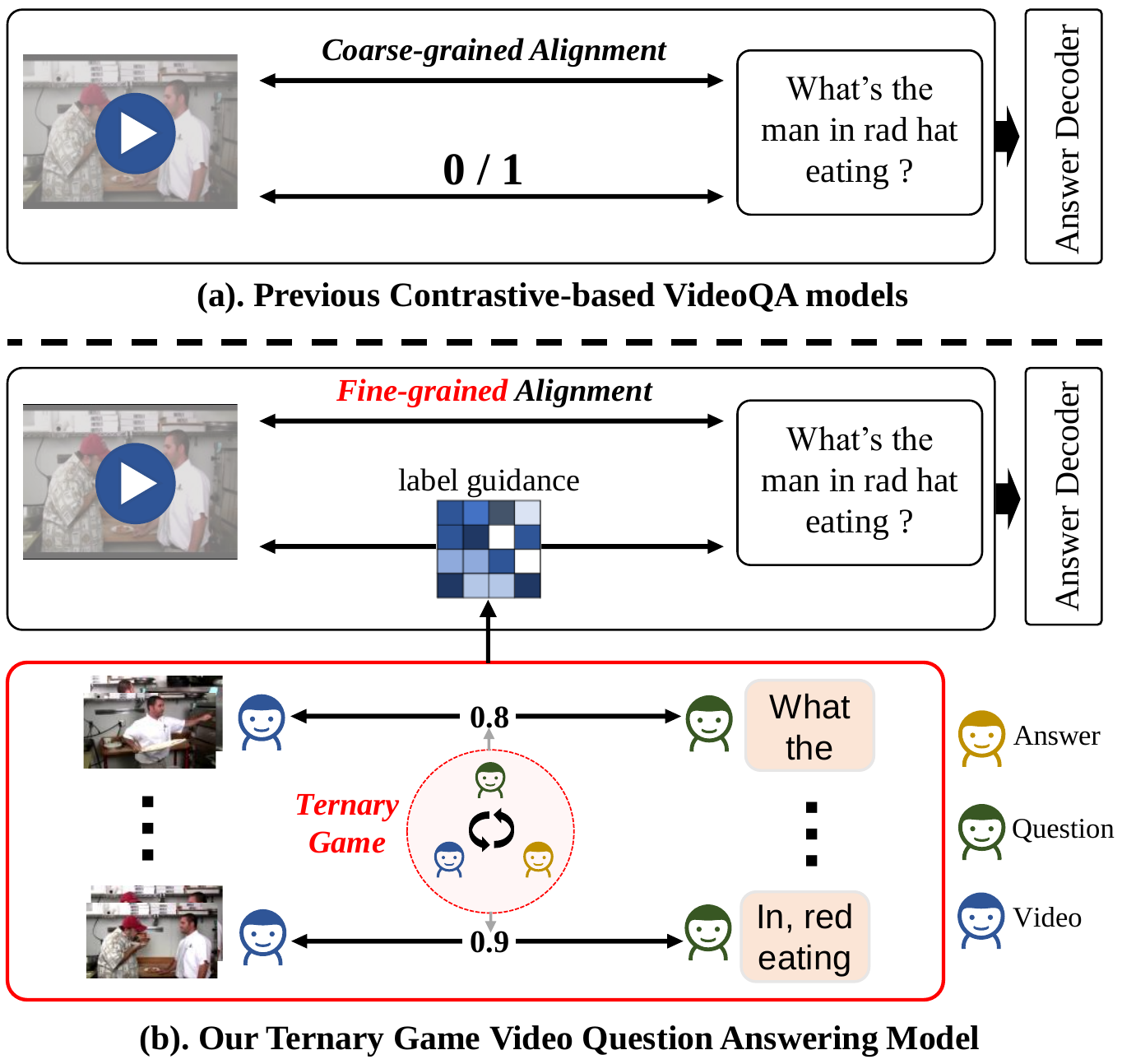}
    \vspace{-1.5em}
    \caption{(a). Contrastive-based VideoQA models only learn a coarse-grained global alignment before the answer decoder. (b).~To achieve fine-grained alignment, we model video, question, and answer as ternary game players and use a VideoQA-specific interaction to generate the 
    label guidance for improvement.
    }
    \label{Motivation}
    \vspace{-1.0em}
\end{figure}

Video question answering(VideoQA)~\cite{wu2017visual,sun2021video} aims to automatically infer the correct answer given a video and a related textual question. Such a multi-modal vision-language task has potential application in a broad range of applications such as vision language navigation in embodied AI~\cite{gu2022vision}, video content retrieval by users' questions~\cite{jin2022expectation,jin2023diffusionret}. Tremendous progress has been made recently in VideoQA, thanks to advances in vision language pre-training and development in model architecture. However, for a VideoQA task, due to the intrinsic property of the visual data, we typically learn from datasets with the \textit{long sequence frames}, which consist of various visual appearances and rich motion information.


The visual data's long sequence property introduces many challenges for multi-modal reasoning in the wild, as a deep learning model has to simultaneously cope with multi-modal representation learning, visual-linguistic alignment, and answer prediction~\cite{li2022invariant,li2022toward}. The naive method would require high-quality annotated data and are typically data-hungry, and remains challenging to achieve accurate visual-linguistic alignment.


Early works on VideoQA focus on developing specific architectures~\cite{jiang2020reasoning,li2021align,qian2022locate} to align the visual representation and the linguistic question, which require sophisticated and heuristic design. More recent efforts\cite{lei2021less,yang2022learning} aims to learn a VideoQA model with contrastive learning by leveraging the power of large-scale dataset, which rarely explore the fine-grained visual-linguistic alignment (shown in Figure~\ref{Motivation}(a)). This severely limits its modeling capacity and generalization ability in answer prediction. Thus, we raise a question: \textit{can we achieve the accurate and robust fine-grained alignment in a data-efficient manner for VideoQA?}

To answer this question, we need to tackle the acquisition of the accurate annotation of fine-grained alignment between question semantics and video clips. However, it is prohibitive to collect the manual annotation due to the mega-scale of the video amount. A promising idea is to automatically generate the alignment annotation without labor-intensive effort.

Toward this goal, we focus on incorporating visual-linguistic alignment into the contrastive learning framework and propose an annotation generator of the fine-grained alignment~(FAG) based on the \textit{multi-player game theory}~\cite{makingsynthesis}. Specifically, we introduce the interaction strategy for annotation generator construction, where the \textit{(video, question, answer)} are treated as ternary game players, and the multi-player game theory could mathematically simulate the pairwise annotation between video clips and question semantics (illustrated in Figure\ref{Motivation}(b)).

In this work, we first carefully design an interaction strategy to tailor the characteristics of VideoQA. Intuitively, if there exists a strong semantic correspondence between the \textit{video player} and the \textit{question player}, and these two players both have a large contribution to the answer, the coalition between them will be strengthened in our framework. Equipped with the annotation generator, we are able to model the fine-grained visual-linguistic alignment with additional supervision signals. To further improve the alignment efficiency, we also explore the multi-modal token reduction strategy for VideoQA. We thoroughly investigate different reduction methods and develop a clustering-based token merge module in the end. Our total framework is named Ternary Game VideoQA~(TG-VQA).

We conduct extensive experiments to validate our model on three VideoQA datasets, including MSVD-QA, MSRVTT-QA, and ActivityNet-QA. The empirical results and ablative studies show our method consistently achieves significant improvements(more than 5\%) on all benchmarks. The annotation generator built from the ternary game also significantly improves the model convergence and data efficiency, which makes our TG-VQA competitive or superior compared with most pre-trained models learned from millions of video data. The main contributions are as follows:


\begin{itemize}
    \item To the best of our knowledge, we are the first to bring game theory into VideoQA. Utilizing game theory's ability to simulate the video-question token relations, our game theory-based annotation generator helps the VideoQA task achieve fine-grained alignment. 

    \item Our alignment label generator is built from the ternary game. For the characteristics of the VideoQA task, the ternary Game models video, question, and answer as ternary game players. The ternary game values the video-question pair’s alignment possibility and their contribution to the answer.
    
    \item We achieve new SoTA results on both short-term and long-term VideoQA datasets, verifying the effectiveness and generalization ability. Without the pretraining stage, our TG-VQA also outperforms most VideoQA pre-trained models.

\end{itemize}
\section{Related Works}
\subsection{Video Question Answering}
The video question answering~(VideoQA) task~\cite{zhong2022video} requires models to analyze the complex semantic correlation between the video and the question. The VideoQA task has two main-stream models: (1). Hierarchical cross-attention-based models. (2). Contrastive learning-based models. Hierarchical cross-attention models~\cite{xu2017video,li2019beyond,li2022joint,peng2022multilevel,cai2021feature,fan2019heterogeneous,ye2023fits} design Spatio-temporal attention structures to fusion the video and text features. 
Several recent models establish the effective alignment stage~\cite{li2021align,xiao2021next} for VideoQA. \cite{jiang2020reasoning} constructs the video clips and text entities into heterogeneous graphs to achieve fine-grained alignment. \cite{li2022invariant} establish the video-question alignment using invariant grounding. \cite{qian2022locate} uses a locator to align the question with video segments. However, these alignment strategies cannot apply in contrastive learning frameworks due to their hierarchical attention structure.

Contrastive learning-based VideoQA models~\cite{lei2021less,kim2020modality,piergiovanni2022video} use contrastive loss for cross-modality explicit alignment and fusion. However, lacking fine-grained alignment annotations, they suffer from slow convergence, requiring massive video data~\cite{bain2021frozen,yang2022learning,huang2021multilingual,li2020hero} for pretraining. Therefore, we establish a ternary game-based contrastive learning VideoQA model, using game-theoretic interaction~\cite{kita1999merging} to generate the fine-grained alignment annotations.



\subsection{Game Theoretic Interaction}
The game-theoretic interaction~\cite{makingsynthesis,ferguson2020course} consists of a set of players with a revenue function. The revenue function maps each team of players to a real number which indicates the payoff obtained by all players working together to complete the task. The core of game-theoretic interaction is to allocate different payoffs to game individuals fairly and reasonably. There are several interaction strategies including core interaction~\cite{jeukenne1977optical}, shapley interaction~\cite{sun2020random} and banzhaf interaction~\cite{marichal2011weighted}. The game-theoretic interaction has multiple applications in different fields~\cite{aflalo2022vl,datta2016algorithmic,tang2021discovering,tang2023learning,li2023multi}. Recently, LOUPE~\cite{li2022fine} uses two-player interaction as a vision-language pre-training task. In this paper, we design a new framework of ternary game interaction strategy for the VideoQA task.

\begin{figure*}  
    \centering 
    \includegraphics[width=1.0\textwidth]{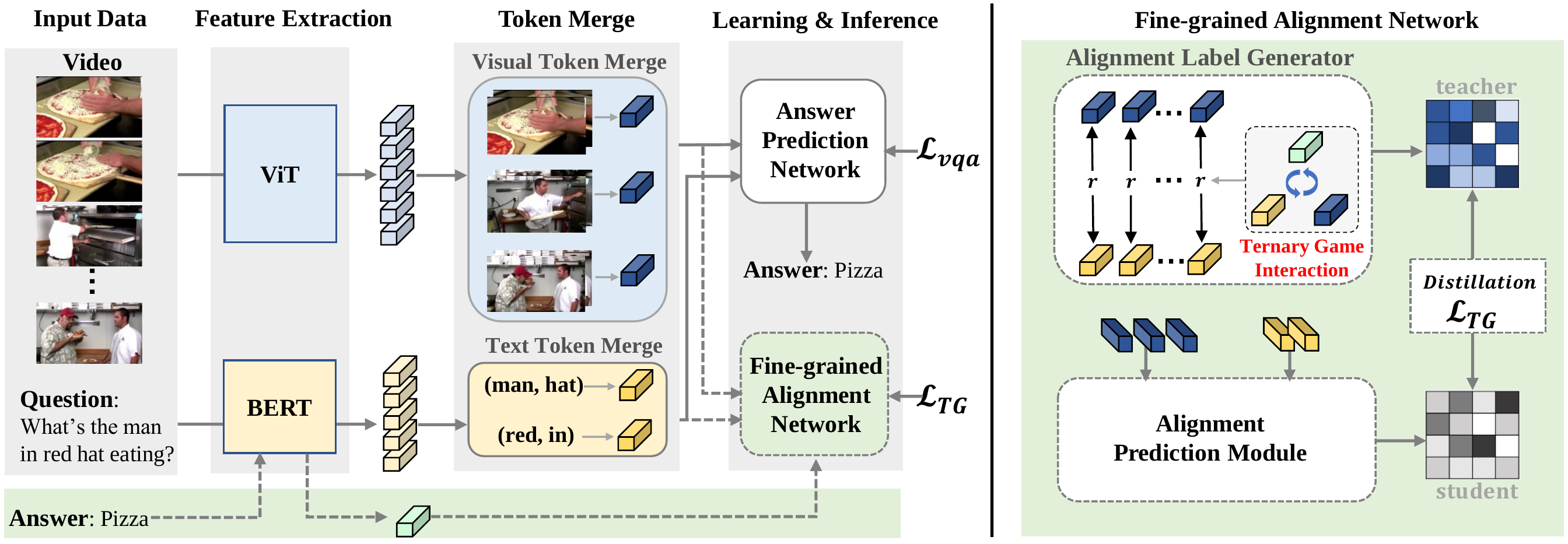}
    \vspace{-0.08in}
    \caption{
    \textbf{The overall framework of our TG-VQA. Left)} We first use a \textit{dual-stream transformer-based encoder} to extract the feature representation for the visual token and question token. We then introduce a \textit{token merge network} to reduce the redundancy of the token and improve the efficiency for visual-linguistic alignment learning. Next, we use the \textit{answer prediction network} to generate the answer for the input video-question pair. Moreover, we develop the \textit{fine-grained alignment network} to explicitly align the visual token and question token at a fine-grained level. \textbf{Right)} The fine-grained alignment network consists of a \textit{alignment label generator}, which is built from the ternary game interaction~(video, question, answer ternary), and an alignment prediction module. We take the similarity matrix produced by the generator as the teacher and distill the fine-grained alignment knowledge from the ternary game interaction to the student. This extra supervision signals improve the consistency between the visual representation and linguistic representation and benefit the multi-modal reasoning in answer prediction.  Notice that structures with \textcolor[RGB]{57,181,74}{green background} and \textcolor{gray}{gray dotted line} are auxiliary components, only used in the training process.
    }
    \label{main_model} 
    \vspace{-1.0em}
\end{figure*}

\section{Preliminary of VideoQA and Game theory}
In this section, we first introduce the problem setting of video question answering in Sec.{\color{red}\ref{problem_setting}}, then briefly present the background for the multi-player game theory(Sec.{\color{red}\ref{preknowledge_game_theory}}), which is utilized in our proposed alignment label generator.

\subsection{Problem Setting of VideoQA}
\label{problem_setting}
Given a video clip ${\textbf{V}}$ and a text-form query ${\textbf{Q}}$, the VideoQA task aims to predict the correct answer $\hat{a}$ from the answer space ${\mathbf{A}}$. For the close-set type of question, ${\mathbf{A}}$ is a fix-size answering option list. For open-ended and multi-choice kinds of questions, ${\mathbf{A}}$ comprises the group of pre-defined answers and a list of candidate answering options. Generally, We formulate the VideoQA task as follows.
\begin{gather}
    \hat{a} = \mathop{argmax}\limits_{a \in {\mathbf{A}}} \mathcal{F}_{\theta}(a | \mathbf{Q}, \mathbf{V})
\end{gather}
where $\theta$ represents the trainable parameters group, $\mathcal{F}_{\theta}$ represents the modeling function.

\subsection{Introduction of Game Theory}
\label{preknowledge_game_theory}
Toward the goal of achieving fine-grained alignment between video and question, we propose to leverage multi-player game theory to construct an alignment label generator. We aim to obtain the semantic relationship between visual tokens and question tokens, and their contribution to the answer. And the game theory targets generating an appropriate coalition construction strategy for multiple players. Thus, we propose to introduce game theory to help the alignment label generation by considering the (video, question, and answer) as players. 

Specifically, the multi-player game theory typically consists of (a) a set of players $\Gamma = (\mathcal{P}, \mathcal{R})$ consists a set of players $\mathcal{P}=\{1, 2, ..., n\}$, and (b) a revenue function $\mathcal{R}(\mathcal{P})$. $\mathcal{R}$ maps each team of players to a real score, which indicates the payoff obtained by all players working together to complete the task. The key step of the game theory is to measure how much gain is obtained, and how to allocate the total gain fairly.



\paragraph{Interaction Strategy} In the multi-player game process, there are various different interaction strategies available, such as \textit{Core interaction}~\cite{jeukenne1977optical}, \textit{Shapley interaction}~\cite{sun2020random}, \textit{Banzhaf interaction}~\cite{marichal2011weighted}. Here we choose the Banzhaf interaction due to its balance of computational complexity and precision. 
Formally, given a coalition $\{i, j\} \in \mathcal{P}$, the Banzhaf interaction $\mathcal{B}(\{i, j\})$ for the player $\{i, j\}$ is defined as:

\begin{gather}
   \mathcal{B}(\{i, j\}) = \sum_{C\in \mathcal{P}\ \{i, j\}}p(C)[\mathcal{R}(C\cup\{i, j\}) + \mathcal{R}(C) \\
    - \mathcal{R}(C\cup\{i\}) - \mathcal{R}(C\cup\{j\})], \nonumber
\label{eq_banzhaf}
\end{gather}
where $\mathcal{P} \ \{i,j\}$ represents removing $\{i,j\}$ from $\mathcal{P}$, $C$ stands for the coalition. $p(C)$ is $\frac{1}{2^{n-2}}$, the possibility for $C$ being sampled.

Intuitively, $\mathcal{B}(\{i, j\})$ reflects the tendency of interactions inside $\{i, j\}$. The higher value of $\mathcal{B}(\{i, j\})$ indicates that player $i$ and player $j$ cooperate closely with each other. For VideoQA, we take the matrix $\mathcal{B}$ as the alignment label annotation by changing the player definition and using a VideoQA-specific interaction strategy. We will start with a
detailed description of our model architecture below.





\section{Method}
Our model consists of four main submodules: (1) a \textbf{backbone network} for generating feature representations 
of the video and question~(Sec.{\color{red}\ref{representation_extraction}}). (2) a \textbf{token merge network} for reducing the visual and question token number~(Sec.{\color{red}\ref{representation_extraction}}). (3) a \textbf{fine-grained alignment network} for establishing the fine-grained visual-linguistic alignment in VideoQA~(Sec.{\color{red}\ref{alignement_network}}). (4) an VideoQA \textbf{answer prediction network} for generating the answer for VideoQA~(Sec.{\color{red}\ref{answer_prediction}}). We finally detail the training objectives and the inference pipeline in Sec.{\color{red}\ref{training_inference}}. The overview of our proposed Ternary Game VideoQA~(TG-VQA) model is illustrated in Figure\ref{main_model}.


\begin{figure}[tbp]
    \includegraphics[width=0.48\textwidth]{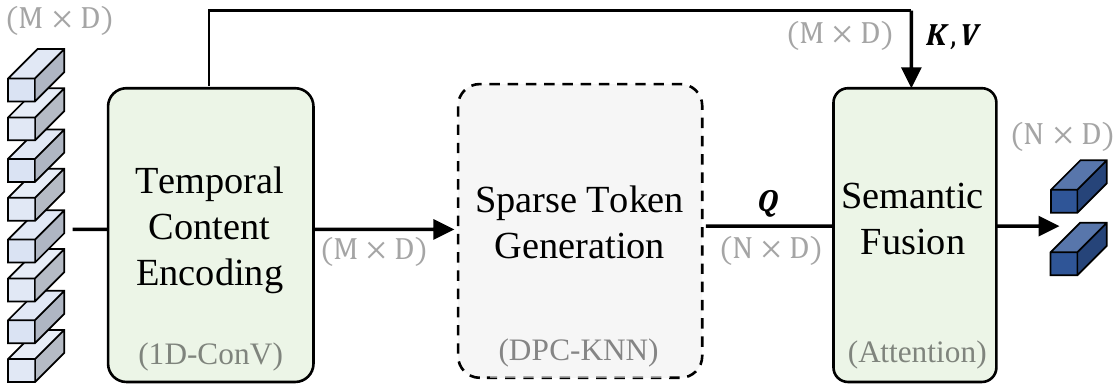}
    \caption{
    We propose the token merge network to reduce the token redundancy. The network contains three modules: the temporal content encoding module, the sparse token generation module, and the semantic fusion module. Here we apply a 1D convolutional layer for temporal encoding,  a DPC-KNN for sparse generation, and an attention layer for semantic fusion.
    }
    \label{token_merge_module}
    \vspace{-0.05in}
\end{figure}

\subsection{Backbone and Token Merge Network}
\label{representation_extraction}
\paragraph{Backbone} We adopt the ViT~\cite{dosovitskiy2020image} and BERT~\cite{devlin2018bert} as the backbone for generating visual representation and textural representation, respectively. Formally, we denote the representation of a video clip as a set of visual tokens $\mathbf{V} = \{\mathbf{v}_{i}|\mathbf{v}_{i} \in \mathbb{R}^{C_v}\}_{i=1}^{N_v}$, where $\mathbf{v}_i$ is one frame feature vector with $C_v$ channel and $N_v$ is the total frame number. For the linguistic representation of question and answer, we first pad them into a fixed length $N_{l}$ sequence and extract their textual feature via a transformer encoder initialized with BERT parameters. We formulate the generated question representation as a set of question tokens: $\mathbf{Q}=\{\mathbf{q}_j|\mathbf{q}_j\in\mathbb{R}^{C_l}\}_{j=1}^{N_l}$, where $\mathbf{q}_j$ is a question token with $C_l$ channel. Similarly, we are able to generate the representation of the answer $\mathbf{A}$ with a text encoder, which is used to construct an alignment label generator.



\paragraph{Token Merge Network} To reduce the redundancy of visual tokens and question tokens, we develop a token merge network and investigate different merge strategies. Typically, the token merge network consists of \textit{(a) temporal context encoding module}, \textit{(b) sparse token generation module} and \textit{(c) semantic fusion module}, which is illustrated in Figure\ref{token_merge_module}. We will focus on the visual token in the remainder of this paragraph for notation clarity. 

Given a sequence of visual tokens $\mathbf{V}$, we first utilize the 1-D convolution layer to encode temporal context efficiently and denote the enhanced visual tokens as $\mathbf{\Tilde{V}}$. Then, we conduct the sparse token generation for the $\mathbf{\Tilde{V}}$ to reduce the number of tokens. Specifically, we investigate several different reduction strategies, like the random initialized sparse token(or called query) and clustering-based sparse token generation. We empirically find that the Density Peaks Clustering based KNN (DPC-KNN)~\cite{rodriguez2014clustering,li2022dynamic} is superior for generating sparse and representative tokens, the ablative studies are shown in Sec.\ref{token_merge_ablation}. we refer the reader to supplementary material for more details of the clustering method. Finally, we apply the cross-attention between the sparse tokens generated from the clustering technique and enhanced visual tokens $\mathbf{\Tilde{V}}$, to further incorporate the semantic context information into the sparse tokens. We denote the spare visual tokens generated from the token merge network as $\mathbf{V}_s \in \mathbb{R}^{C_v\times N_{vs}}$, where $N_{vs}$ is the number of sparse visual tokens and we have $N_{vs}<N_{v}$. Similarly, we are able to produce the spare textual tokens $\mathbf{Q}_s \in \mathbb{R}^{C_l\times N_{ls}}$ from the token merge network. Equipped with spare visual and question tokens, we not only reduce the redundancy in the input data but also improve the fine-grained visual-linguistic alignment efficiency~(more discussion is presented in Sec.~\ref{alignement_network}).

\subsection{Fine-grained Alignment Network}
\label{alignement_network}
Different from the current contrastive learning-based methods, which adopt the coarse-grained visual-linguistic alignment in model optimization, we develop a fine-grained alignment network that serves as an auxiliary network only existing in the training process, to explicitly supervise the model with an automatically generated alignment label.

The main idea is to introduce an alignment label generator to provide a supervision signal for learning a VideoQA model. We rethink the relationship among each item(video, question, and answer) and find that the alignment between visual tokens and question tokens actually reflects the semantic correspondence. Tokens sharing similar semantic meanings tend to simultaneously contribute to the final answer prediction. Thus, we propose to leverage the multi-player game theory to help find tokens with high semantic similarity. The fine-grained alignment network is composed of (a) an alignment label generator constructed with ternary game theory, and (b) an alignment prediction module.


\begin{table*}
    \centering
    \setlength{\tabcolsep}{3.5mm}        
    \begin{tabular}{llrrrr}
        \toprule[1.5pt]
        \# & Model & Initialization & Pretrain Data & MSRVTT-QA & MSVD-QA \\
        \midrule
        \multicolumn{2}{l}{\emph{\textbf{Pretrained}}} & & & \\
        1 & VideoCLIP~\cite{luo2022clip4clip} & S3D+BERT & HowTo100M & 33.8 & 31.8  \\
        2 & ClipBERT~\cite{lei2021less} & ResNet+BERT & HowTo100M & 37.4 & -  \\
        3 & CoMVT~\cite{seo2021look} & S3D+BERT & HowTo100M & 39.5 & 42.6       \\
        4 & VQA-T~\cite{yang2022learning} & S3D+BERT & HowToVQA69M & 41.5 & 46.3  \\
        5 & ALPRO~\cite{li2022align} & ResNet+BERT & Web2M+CC3M & 42.1 & 45.9  \\
        \rowcolor{gray!10} 6 & Co-Tok~\cite{piergiovanni2022video} & K600+T5 & How100M & 45.7 & 48.6 \\
        \rowcolor{gray!10} 7 & FrozenBiLM~\cite{yang2022zero} & CLIP+GPT3 & WebVid10M & 47.0 & 54.8 \\
        \midrule
        \multicolumn{2}{l}{\emph{\textbf{Non-Pretrained}}} & & &\\
        8 & HCRN~\cite{le2020hierarchical} & ResNet+LSTM & None & 35.6 & 35.5      \\
        9 & MHN~\cite{peng2022multilevel} & ResNet+LSTM & None & 38.6 & 40.4     \\
        10 & IGV~\cite{li2022invariant} & ResNet+BERT & None & 38.3 & 40.8       \\
        11 & VQA-T~\cite{yang2022learning} & S3D+BERT & None & 39.6 & 41.2        \\
        12 & CLIP-QA~\cite{radford2021learning} & CLIP+BERT & None & 39.0 & 38.5        \\
        13 & CLIP4clip~\cite{luo2022clip4clip} & CLIP+BERT & None & 40.9 & 39.3        \\
        14 & \textbf{TG-VQA~(ours)} & S3D+BERT & None & 42.7 & 45.5 \\
        15 & \textbf{TG-VQA~(ours)} & CLIP+BERT & None & \textbf{46.3} & \textbf{52.5} \\
        \bottomrule[1.5pt]
    \end{tabular}
    \vspace{-0.05in}
    \caption{Experiments for the MSRVTT-QA and MSVD-QA datasets. We surpass the non-pretrained VideoQA models by a wide margin. Without the large-scale pretraining dataset, our interaction model also surpasses most of the pretrained VideoQA models.}
    \vspace{-0.1in}
    \label{MSRVTT}
\end{table*}

\begin{table}[]
    \centering
    \setlength{\tabcolsep}{2.5mm}        
    \begin{tabular}{llrrr}
        \toprule[1.5pt]
        \# & Model & Initialization & Pretrain & Acc. \\
        \midrule
        \multicolumn{2}{l}{\emph{\textbf{Pretrained}}} & & \\
        1 & VQA-T & S3D+BERT & 69M & 38.9 \\
        2 & LF-VILA & Swin+BERT & 8M & 39.9 \\
        3 & FrozenBiLM & CLIP+GPT3 & 10M & 43.2 \\
        4 & DeST & Swin+BERT & 14M & 46.8 \\
        \midrule
        \multicolumn{2}{l}{\emph{\textbf{Non-Pretrained}}} & & \\
        5 & LocAns & C3D+BERT & - & 36.1 \\ 
        6 & VQA-T & S3D+BERT & - & 36.8  \\
        7 & \textbf{TG-VQA~(ours)} & CLIP+BERT & - & \textbf{48.3} \\
        \bottomrule[1.5pt]
    \end{tabular}
    \vspace{-0.05in}
    \caption{Experiments of the long-term dataset, ActivityNet-QA. Our method surpasses all pretrained and non-pretrained VideoQA models.}
    \label{ActivityNet}
    \vspace{-1.0em}
\end{table}

\paragraph{Alignment Label Generator} Given the spare visual tokens $\mathbf{V}_s$ and sparse question tokens $\mathbf{Q}_s$, we consider video, question and the answer $\mathbf{A}$ as game players, which means $\mathcal{N} = \mathbf{V}_s \cup \mathbf{Q}_s \cup \mathbf{A} $. Intuitively, if a visual token has strong semantic correspondence with a question token, then they tend to cooperate with each other and contribute to the final answer. We now present the \textbf{ternary game interaction} strategy used in our work. For simplicity, we apply Banzhaf~\cite{marichal2011weighted} interaction for the ternary game. Concretely, a task-specific revenue function is required for interaction, and we need to consider the fine-grained visual-linguistic alignments as well as the token pair's contribution to the final answer. Thus, the revenue function $\mathcal{R}$ should satisfy the following criteria(we omit the subscript $s$ of $\mathbf{V}_s, \mathbf{Q}_s$ for clarity in the following paragraph):
\begin{itemize}
    \item $\mathcal{R}(\mathbf{v}_i, \mathbf{q}_j)$ benefits from the semantic similarity between \textit{video} and \textit{question}.
    \item $\mathcal{R}(\mathbf{v}_i, \mathbf{q}_j)$ benefits from the semantic similarity between \textit{the target answer representation $\mathbf{A}$} and \textit{the prediction from video-question pair}.
\end{itemize}
Thus, our proposed revenue function is formulated as:
\begin{gather}
    \mathcal{R}(\mathbf{v}_i, \mathbf{q}_j, \mathbf{A}) = \phi(\mathbf{v}_i, \mathbf{q}_j) + \phi(\mathbf{A}, \mathbf{G}(\mathbf{v}_i,\mathbf{q}_j)).
\end{gather}
where $\phi$ is a distance measurement for the semantic similarity, $\mathbf{G}$ is a linear layer to project the  concatenation of $\mathbf{v}_i$ and $\mathbf{q}_j$ into the answer representation space. 

Then we apply $\mathcal{R}$ to Eq.\ref{eq_banzhaf} for the Banzhaf interaction.
However, we find brute-force computation of Eq.\ref{eq_banzhaf} is time-consuming. To speed up the interaction calculation process, we propose a deep learning-based idea by using a tiny convolutional network to predict the revenue function. Specifically, so we first calculate 1000 samples' guidance matrix. We can use these matrices as data samples to learn a tiny model with short epochs. We take such a model as the alignment label generator to generate the guidance matrix $\mathcal({V}, {Q})$. 



\paragraph{Alignment Prediction Module} We adopt the contrastive learning-based framework to optimize the VideoQA model similar to~\cite {lei2021less}. Differently, we introduce the explicit supervision signal for fine-grained visual-linguistic alignment. We first generate the alignment prediction $\mathbf{{R}}(\mathbf{V}_s, \mathbf{Q}_s))$ between visual tokens and linguistic tokens by computing their similarity. Then, we regard the guidance matrix generated from the alignment label generator as the teacher, and the alignment prediction as the student. The model is optimized by minimizing the Kullback-Leibler (KL) divergence between teacher and student. The ternary game loss $\mathcal{L}_{TG}$ is 
\begin{gather}
    \mathcal{L}_{TG} = \mathbb{E}_{\mathbf{V}_s,\mathbf{Q}_s}[ \text{KL}(\mathcal{R}(\mathbf{V}_s, \mathbf{Q}_s), \mathbf{{R}}(\mathbf{V}_s, \mathbf{Q}_s)) ] 
\end{gather}
with such a distillation process, the model is expected to learn the multi-modal representation with rich semantic information and fine-grained visual-linguistic alignment.



\subsection{Answer Prediction Network}
\label{answer_prediction}
Due to the established visual-linguistic fine-grained alignment benefit from the FAN, we are able to adopt a simplified answer prediction network, without the need for sophisticated multi-modal fusion/reasoning stages like many previous VideoQA models. 

Specifically, given the sparse visual tokens $\mathbf{V}_s$ and sparse textual tokens $\mathbf{Q}_s$ generated from the token merge network, we first predict the token-level fusion weight by applying the non-linear projection(typically using linear layer + sigmoid function) on each token of $\mathbf{V}_s$ and $\mathbf{Q}_s$. We then obtain the global representation for each modality by conducting weighted sum in $\mathbf{V}_s$ and $\mathbf{Q}_s$, respectively. We denote the generated global feature vectors as $\mathbf{v}_o$ and $\mathbf{q}_o$. Next, we concatenate those two vectors followed by an MLP to predict the answer $logits$. We use cross-entropy loss between $logits$ and ground truth answer to supervise the whole framework.


\subsection{Training and Inference}
\label{training_inference}
Combining the $\mathcal{L}_{TG}$ from the ternary game module and the cross-entropy loss from the vqa prediction module, the overall loss of our model is the weighted sum of both parts:
\begin{gather}
    \mathcal{L} = \mathcal{L}_{vqa} + \alpha\mathcal{L}_{TG}.
\end{gather}
$\alpha$ is the trade-off hyper-parameter for the ternary game.  
As shown in Figure.~\ref{main_model}, the structures with green
backgrounds and Dotted lines are auxiliary components, which only appear in training. During the inference process, TG-VQA only activates the answer prediction network.

\begin{table*}
\begin{minipage}[c]{0.25\textwidth}
\centering
\renewcommand{\arraystretch}{1.35}  
\resizebox{1.0\linewidth}{!}{
    \begin{tabular}{c|c}
        \toprule[1.5pt]
         Model & Acc. \\
         \midrule
         Baseline & 39.0       \\
         Baseline+FAN & 43.1    \\
         Baseline+FAN+TM & \textbf{46.3} \\ 
         \bottomrule[1.5pt]
    \end{tabular}
}
\vspace{-0.08in}
\caption{Ablation for our fine-grained alignment network (FAN) and the token merge module (TM). Both modules have benefits.}
\label{Total_ablation}
\vspace{-1.0em}
\end{minipage}
\hfill
\begin{minipage}[c]{0.4\textwidth}
\centering
\resizebox{1.0\linewidth}{!}{
    \begin{tabular}{lc|r}
        \toprule[1.5pt]
         Alignment & Strategy & Acc. \\
         \midrule
         Coarse-grained & - & 39.0\\
         \midrule
         \multirow{2}{*}{Fine-grained} & Fully-connect & 43.1\\
          & Ternary Game  & 44.5\\
         \bottomrule[1.5pt]
    \end{tabular}
}
\vspace{-0.08in}
\caption{The ablation for the alignment strategy. Both fine-grained alignment methods surpass the coarse-grained baseline. Our ternary game strategy outperforms the intuitively fully-connected strategy.}
\label{Alignment_strategy}
\vspace{-1.0em}
\end{minipage}
\hfill
\begin{minipage}[c]{0.3\textwidth}
\centering

\resizebox{0.8\textwidth}{!}{
    \begin{tabular}{cc|c}
        \toprule[1.5pt]
         TM & Strategy & Acc. \\
         \midrule
         \XSolidBrush & - & 44.5 \\
         \midrule
         \multirow{2}{*}{\Checkmark} & Random & 41.5\\
         & Temporal & 43.7    \\
         & DPC-KNN & \textbf{46.3} \\ 
         \bottomrule[1.5pt]
    \end{tabular}
}
\vspace{-0.08in}
\caption{Ablation for the cluster strategy in the token merge module(TM). Random cluster and temperal cluster have negative impacts, while DPC-KNN performs well. }
\label{Cluster_strategy}
\vspace{-1.0em}
\end{minipage}
\end{table*}

\begin{figure*}
    \centering
    \setlength{\abovecaptionskip}{0.28cm}
    \includegraphics[width=1.0\linewidth]{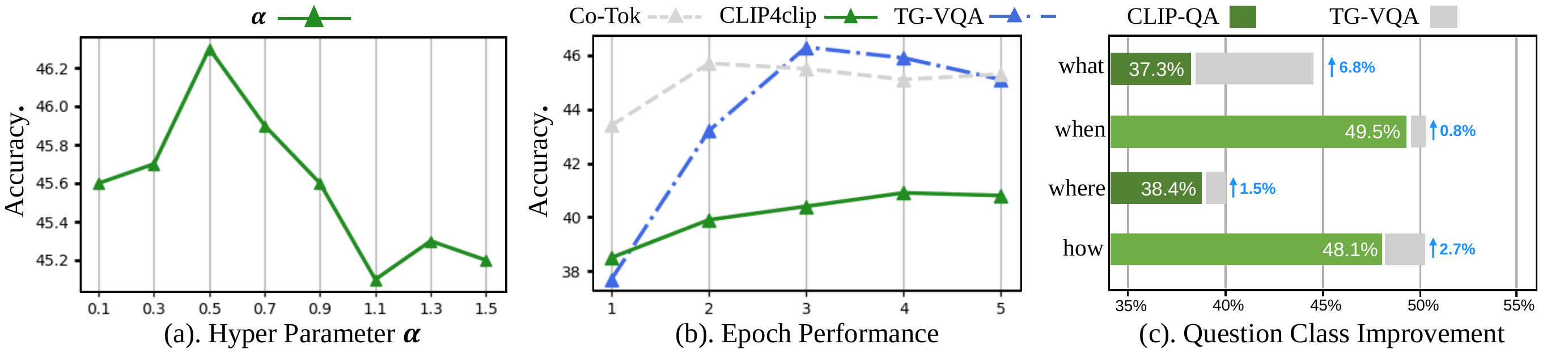}
    \vspace{-1.0em}
    \caption{(a). The ablation study for hyperparameter $\alpha$, the weight of $\mathcal{L}_{TG}$. Our TG-VQA performs best when $\alpha=0.5$. (b). The ablation study for epoch performance between different VideoQA models. With our ternary game module, the model converges faster. (c). The performance analysis for the question category. Our ternary game module significantly improves ``what'' and ``how'' question correctness.}
    \label{Graph}
    \vspace{-1.0em}
\end{figure*}

\section{Experiments}

\subsection{Datasets}
We select multiple VideoQA datasets to comprehensively evaluate the effectiveness of our method on different-length videos. Following the VQA-T~\cite{yang2022learning} setting, we choose two short video datasets ~(MSVD-QA, MSRVTT-QA) and one long video dataset~(ActivityNet-QA) as our evaluation benchmarks.
\textbf{MSVD-QA}~\cite{xu2017video} comprises 1,970 short clips and 50,505 question-answer pairs. The clip's average length is 10 seconds and the questions are divided into five question categories: what, who, how, when, and where. All of them are open-ended.
\textbf{MSRVTT-QA}~\cite{xu2017video} comprises 10K videos and 243K question-answer pairs. The question types are similar to what is included in the MSVD-QA dataset. However, the scenario of the video is more complex, with a longer duration of 10-30 seconds.
\textbf{ActivityNet-QA}~\cite{yu2019activitynet} is a Human annotated and large scale VideoQA dataset. The dataset consists of 58,000 QA pairs on 5,800 complex long web videos derived from the popular ActivityNet dataset. The average video length of ActivityNet-QA is 180 seconds, which is much longer than MSRVTT-QA and MSVD-QA.

\subsection{Experimental Results}
We select the most recent pretrained and non-pretrained VideoQA models for comparison. Table.~\ref{MSRVTT} shows the experimental results on the MSRVTT-QA dataset and MSVD-QA dataset. Compared with the non-pretrained VideoQA models, our method achieves sustainable improvements, 5.4\% on MSRVTT-QA and 11.3\% on MSVD-QA. Without millions of video-text pretraining data, our method also surpasses most of the pretrained VideoQA models. Table.~\ref{ActivityNet} shows the experimental results on the long-term VideoQA dataset, ActivityNet-QA. Our model achieves 48.3\%, surpassing all pretrained and non-pretrained VideoQA models.

\subsection{Ablation Studies}
We first explore each module's contribute to the overall TG-VQA performance. As shown in Table.~\ref{Total_ablation}. Both our fine-grained alignment network~(FAN) and the token merge module~(TM) benefit the performanc of TG-VQA. More ablation cases are presented below.

\subsubsection{Effectiveness of Alignment Strategies.}
Table.~\ref{Alignment_strategy} explore the benefits of various alignment strategies for the VideoQA task. 
Comparing line~1 and line~2, fine-grained alignment in VideoQA yields 4.1\% benefits from coarse-grained alignment. 
For fine-grained alignment strategies, our ternary game alignment outperforms the intuitively fully-connected alignment by 1.4\%.

\subsubsection{Effectiveness of Token Merge Module. }\label{token_merge_ablation}
Our token merge module cluster the video and question tokens aiming to reduce the token amount for later interaction. In Table.~\ref{Cluster_strategy}, we make several attempts at the clustering strategies. ``Rand Init'' represents randomly initializing the cluster center. ``Temporal'' represents clustering the tokens by temporal or sequence order. ``DPC-KNN'' represents our DPC-KNN clustering strategy. Compared with the line~1 using Banzhaf interaction with no merge module, both random cluster and temperal cluster have a negative impact on the model performance due to the Lack of semantic correlation. Meanwhile, our DPC-KNN clustering strategy adaptively merges tokens under the guidance of semantic similarity, surpassing other clustering strategies.

\renewcommand{\dblfloatpagefraction}{.8}
\begin{figure*}[htbp]  
\vspace{-0.1in}
    \centering 
    \includegraphics[width=1.0\textwidth]{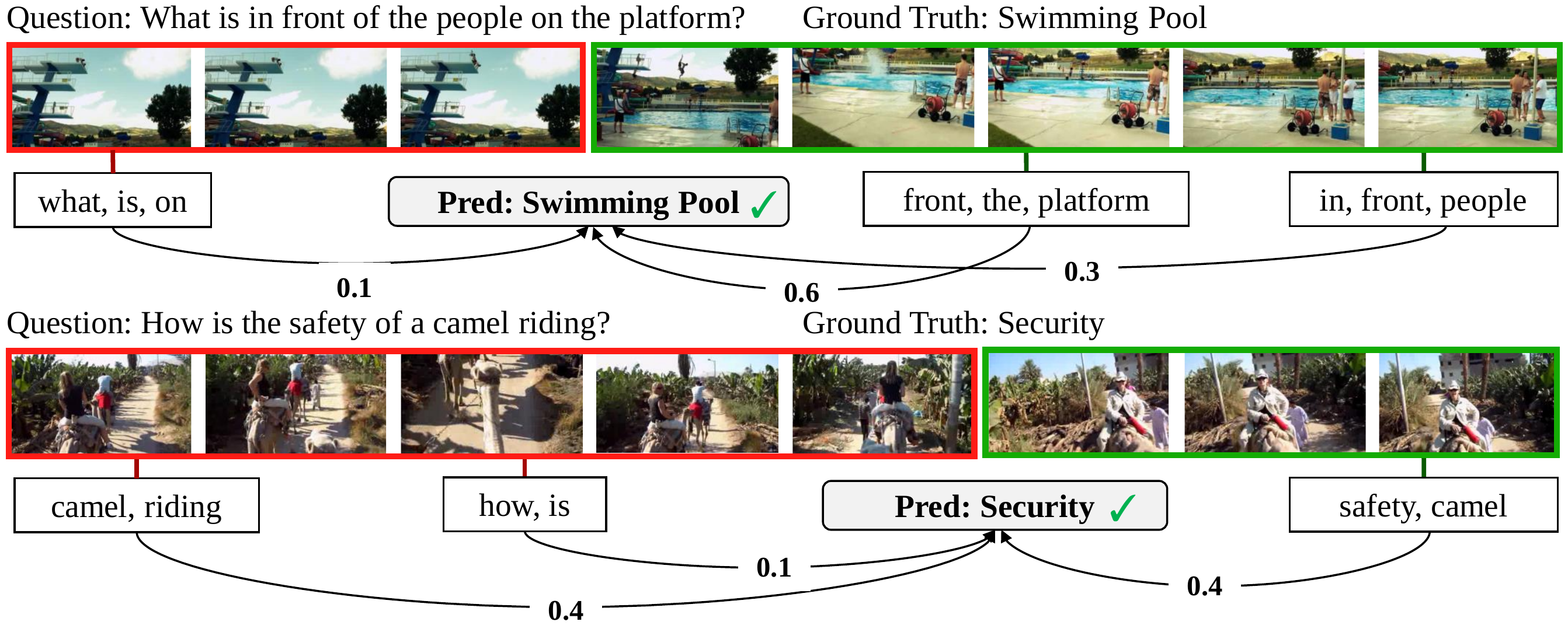}
    \vspace{-2.0em}
    \caption{
    The case visualization. We visualize the most possible alignment pairs between video centers and question text centers. The arrow curves are the visualization of different video-question pairs' contributions to the answer. The visualization shows that our Ternary Game Interaction can correctly align the video-question pair and evaluate their contributions to the final answer.
    }
    \label{visualization} 
    \vspace{-1.0em}
\end{figure*}

\subsubsection{Impact of Encoders' Initial Parameters }
We list all VideoQA initial parameter combinations in Table.~\ref{MSRVTT} row~3. Non-pretrained VideoQA models tend to use ResNet, S3D~\cite{xie2018rethinking} or CLIP as video encoder parameters, while using BERT as text encoder parameters. Pretrained models also adapt huge language models as the encoder, including T5~\cite{raffel2020exploring} and GPT~\cite{radford2018improving}. For a fair comparison, we apply the most common CLIP+BERT and S3D+BERT combinations. Shown in Table.~\ref{MSRVTT}, our TG-VQA model with S3D+BERT initialization surpasses other non-Pretrained VideoQA models with S3D+BERT. Using CLIP+BERT initialization, our TG-VQA model outperforms others by 5.4\% in MSRVTT-QA and 11.7\% in MSVD-QA. 
Due to FrozenBiLM's large computation from GPT3 and WebVid10M, we don't compare with FrozenBiLM.

\subsubsection{Hyper Parameters in Train Objective}
In order to explore the effect of the ternary game loss $\mathcal{L}_{TG}$'s hyperparameter on the performance of the model, we train our TG-VQA on the MSRVTT-QA dataset with hyperparameter $\alpha$ from 0.1 to 1.5. Shown in Figure\ref{Graph}~(a), the model performance fluctuates in a range $[45.1, ~46.3]$. When $\alpha = 0.5$, our TG-VQA performs best with 46.3 accuracy.

\subsubsection{Epoch Analysis}
To illustrate our ternary game's ability to accelerate the model's convergence process with limited data, we visualize the epoch performance for our TG-VQA model and the CLIP4clip~(non-pretrained VideoQA model without the ternary game) on MSRVTT. In Figure\ref{Graph}~(b), both CLIP4clip and TG-VQA apply the same encoder initialization. With the fine-grained alignment network, our TG-VQA converges faster and better than CLIP4clip. We also visualize the epoch curve of the Co-Tok~(a pretrained VideoQA model). With limited data, our TG-VQA surpasses Co-Tok on epoch~3, demonstrating our data-efficency.

\subsubsection{Question Category Performance Analysis}
We visualize the model's performance on the Top-4 question categories on the MSRVTT-QA dataset. As shown in Figure\ref{Graph}~(b), with the addition of our ternary game module, the model significantly improves ``what'' and ``how'' question types performance, which attributes to the fine-grained alignment brought by the ternary game module.

\vspace{-0.5em}
\subsection{Case Visualization}
Figure~\ref{visualization} is the case visualization from the ActivityNet-QA dataset. 
For visualizing the cluster results, we cluster the video clips into two centers and the question tokens into three centers. Both cases show semantic similarity within the same centers.
For visualizing the alignment results, Figure~\ref{visualization} shows the top-1 alignment pairs between the video center and the question center. The alignment results conform to the semantic consistency. 
For visualizing the contribution to the prediction answer, 
both cases illustrate that when a video-question pair is unlikely to be the answer, its contribution score is rather low~(0.1). The second case illustrates that when multiple video-question pairs are similar to the answer, their contribution scores also tend to be the same. 
The visualization of two cases demonstrates the interpretability of our model.


\vspace{-0.5em}
\section{Conclusion}

In this paper, we study the fine-grained alignment in the VideoQA task. We innovatively model the VideoQA task as a ternary game process between video, question, and answer. We design a VideoQA-specific interaction strategy to simulate the alignment relationship. Experiments show the effectiveness, generalization, and data efficiency of our model.

\noindent \textbf{Acknowledgements.} This work was supported in part by the National Key R\&D Program of China (No. 2022ZD0118201), Natural Science Foundation of China (No. 61972217, 32071459, 62176249, 62006133, 62271465), and the Natural Science Foundation of Guangdong Province in China (No. 2019B1515120049).

\clearpage

\bibliographystyle{named}
\bibliography{ijcai23}

\begin{thebibliography}{}

\bibitem[\protect\citeauthoryear{Aflalo \bgroup \em et al.\egroup
  }{2022}]{aflalo2022vl}
Estelle Aflalo, Meng Du, Shao-Yen Tseng, Yongfei Liu, Chenfei Wu, Nan Duan, and
  Vasudev Lal.
\newblock Vl-interpret: An interactive visualization tool for interpreting
  vision-language transformers.
\newblock In {\em Proceedings of the IEEE/CVF Conference on Computer Vision and
  Pattern Recognition}, pages 21406--21415, 2022.

\bibitem[\protect\citeauthoryear{Bain \bgroup \em et al.\egroup
  }{2021}]{bain2021frozen}
Max Bain, Arsha Nagrani, G{\"u}l Varol, and Andrew Zisserman.
\newblock Frozen in time: A joint video and image encoder for end-to-end
  retrieval.
\newblock In {\em Proceedings of the IEEE/CVF International Conference on
  Computer Vision}, pages 1728--1738, 2021.

\bibitem[\protect\citeauthoryear{Cai \bgroup \em et al.\egroup
  }{2021}]{cai2021feature}
Jiayin Cai, Chun Yuan, Cheng Shi, Lei Li, Yangyang Cheng, and Ying Shan.
\newblock Feature augmented memory with global attention network for videoqa.
\newblock In {\em Proceedings of the Twenty-Ninth International Conference on
  International Joint Conferences on Artificial Intelligence}, pages 998--1004,
  2021.

\bibitem[\protect\citeauthoryear{Datta \bgroup \em et al.\egroup
  }{2016}]{datta2016algorithmic}
Anupam Datta, Shayak Sen, and Yair Zick.
\newblock Algorithmic transparency via quantitative input influence: Theory and
  experiments with learning systems.
\newblock In {\em 2016 IEEE symposium on security and privacy (SP)}, pages
  598--617. IEEE, 2016.

\bibitem[\protect\citeauthoryear{Devlin \bgroup \em et al.\egroup
  }{2018}]{devlin2018bert}
Jacob Devlin, Ming-Wei Chang, Kenton Lee, and Kristina Toutanova.
\newblock Bert: Pre-training of deep bidirectional transformers for language
  understanding.
\newblock {\em arXiv preprint arXiv:1810.04805}, 2018.

\bibitem[\protect\citeauthoryear{Dosovitskiy \bgroup \em et al.\egroup
  }{2020}]{dosovitskiy2020image}
Alexey Dosovitskiy, Lucas Beyer, Alexander Kolesnikov, Dirk Weissenborn,
  Xiaohua Zhai, Thomas Unterthiner, Mostafa Dehghani, Matthias Minderer, Georg
  Heigold, Sylvain Gelly, et~al.
\newblock An image is worth 16x16 words: Transformers for image recognition at
  scale.
\newblock {\em arXiv preprint arXiv:2010.11929}, 2020.

\bibitem[\protect\citeauthoryear{Fan \bgroup \em et al.\egroup
  }{2019}]{fan2019heterogeneous}
Chenyou Fan, Xiaofan Zhang, Shu Zhang, Wensheng Wang, Chi Zhang, and Heng
  Huang.
\newblock Heterogeneous memory enhanced multimodal attention model for video
  question answering.
\newblock In {\em Proceedings of the IEEE/CVF conference on computer vision and
  pattern recognition}, pages 1999--2007, 2019.

\bibitem[\protect\citeauthoryear{Fang \bgroup \em et al.\egroup
  }{2023}]{tang2023learning}
Yuchen Fang, Zhenggang Tang, Kan Ren, Weiqing Liu, Li~Zhao, Jiang Bian,
  Dongsheng Li, Weinan Zhang, Yong Yu, and Tieyan Liu.
\newblock Learning multi-agent intention-aware communication for optimal
  multi-order execution in finance.
\newblock In {\em Proceedings of the 29th ACM SIGKDD Conference on Knowledge
  Discovery and Data Mining}, 2023.

\bibitem[\protect\citeauthoryear{Ferguson}{2020}]{ferguson2020course}
Thomas~S Ferguson.
\newblock {\em A course in game theory}.
\newblock World Scientific, 2020.

\bibitem[\protect\citeauthoryear{Gu \bgroup \em et al.\egroup
  }{2022}]{gu2022vision}
Jing Gu, Eliana Stefani, Qi~Wu, Jesse Thomason, and Xin~Eric Wang.
\newblock Vision-and-language navigation: A survey of tasks, methods, and
  future directions.
\newblock {\em arXiv preprint arXiv:2203.12667}, 2022.

\bibitem[\protect\citeauthoryear{Huang \bgroup \em et al.\egroup
  }{2021}]{huang2021multilingual}
Po-Yao Huang, Mandela Patrick, Junjie Hu, Graham Neubig, Florian Metze, and
  Alexander Hauptmann.
\newblock Multilingual multimodal pre-training for zero-shot cross-lingual
  transfer of vision-language models.
\newblock {\em arXiv preprint arXiv:2103.08849}, 2021.

\bibitem[\protect\citeauthoryear{Jeukenne \bgroup \em et al.\egroup
  }{1977}]{jeukenne1977optical}
J-P Jeukenne, A~Lejeune, and C~Mahaux.
\newblock Optical-model potential in finite nuclei from reid's hard core
  interaction.
\newblock {\em Physical Review C}, 16(1):80, 1977.

\bibitem[\protect\citeauthoryear{Jiang and Han}{2020}]{jiang2020reasoning}
Pin Jiang and Yahong Han.
\newblock Reasoning with heterogeneous graph alignment for video question
  answering.
\newblock In {\em Proceedings of the AAAI Conference on Artificial
  Intelligence}, 2020.

\bibitem[\protect\citeauthoryear{Jin \bgroup \em et al.\egroup
  }{2022}]{jin2022expectation}
Peng Jin, Jinfa Huang, Fenglin Liu, Xian Wu, Shen Ge, Guoli Song, David
  Clifton, and Jie Chen.
\newblock Expectation-maximization contrastive learning for compact
  video-and-language representations.
\newblock {\em Advances in Neural Information Processing Systems}, 2022.

\bibitem[\protect\citeauthoryear{Jin \bgroup \em et al.\egroup
  }{2023}]{jin2023diffusionret}
Peng Jin, Hao Li, Zesen Cheng, Kehan Li, Xiangyang Ji, Chang Liu, Li~Yuan, and
  Jie Chen.
\newblock Diffusionret: Generative text-video retrieval with diffusion model.
\newblock {\em arXiv preprint arXiv:2303.09867}, 2023.

\bibitem[\protect\citeauthoryear{Kim \bgroup \em et al.\egroup
  }{2020}]{kim2020modality}
Junyeong Kim, Minuk Ma, Trung Pham, Kyungsu Kim, and Chang~D Yoo.
\newblock Modality shifting attention network for multi-modal video question
  answering.
\newblock In {\em Proceedings of the IEEE/CVF conference on computer vision and
  pattern recognition}, pages 10106--10115, 2020.

\bibitem[\protect\citeauthoryear{Kita}{1999}]{kita1999merging}
Hideyuki Kita.
\newblock A merging--giveway interaction model of cars in a merging section: a
  game theoretic analysis.
\newblock {\em Transportation Research Part A: Policy and Practice},
  33(3-4):305--312, 1999.

\bibitem[\protect\citeauthoryear{Le \bgroup \em et al.\egroup
  }{2020}]{le2020hierarchical}
Thao~Minh Le, Vuong Le, Svetha Venkatesh, and Truyen Tran.
\newblock Hierarchical conditional relation networks for video question
  answering.
\newblock In {\em Proceedings of the IEEE/CVF conference on computer vision and
  pattern recognition}, pages 9972--9981, 2020.

\bibitem[\protect\citeauthoryear{Lei \bgroup \em et al.\egroup
  }{2021}]{lei2021less}
Jie Lei, Linjie Li, Luowei Zhou, Zhe Gan, Tamara~L Berg, Mohit Bansal, and
  Jingjing Liu.
\newblock Less is more: Clipbert for video-and-language learning via sparse
  sampling.
\newblock In {\em Proceedings of the IEEE/CVF Conference on Computer Vision and
  Pattern Recognition}, 2021.

\bibitem[\protect\citeauthoryear{Li \bgroup \em et al.\egroup
  }{2019}]{li2019beyond}
Xiangpeng Li, Jingkuan Song, Lianli Gao, Xianglong Liu, Wenbing Huang, Xiangnan
  He, and Chuang Gan.
\newblock Beyond rnns: Positional self-attention with co-attention for video
  question answering.
\newblock In {\em Proceedings of the AAAI Conference on Artificial
  Intelligence}, volume~33, pages 8658--8665, 2019.

\bibitem[\protect\citeauthoryear{Li \bgroup \em et al.\egroup
  }{2020}]{li2020hero}
Linjie Li, Yen-Chun Chen, Yu~Cheng, Zhe Gan, Licheng Yu, and Jingjing Liu.
\newblock Hero: Hierarchical encoder for video+ language omni-representation
  pre-training.
\newblock {\em arXiv preprint arXiv:2005.00200}, 2020.

\bibitem[\protect\citeauthoryear{Li \bgroup \em et al.\egroup
  }{2021}]{li2021align}
Junnan Li, Ramprasaath Selvaraju, Akhilesh Gotmare, Shafiq Joty, Caiming Xiong,
  and Steven Chu~Hong Hoi.
\newblock Align before fuse: Vision and language representation learning with
  momentum distillation.
\newblock {\em NIPS}, 34:9694--9705, 2021.

\bibitem[\protect\citeauthoryear{Li \bgroup \em et al.\egroup
  }{2022a}]{li2022align}
Dongxu Li, Junnan Li, Hongdong Li, Juan~Carlos Niebles, and Steven~CH Hoi.
\newblock Align and prompt: Video-and-language pre-training with entity
  prompts.
\newblock In {\em Proceedings of the IEEE/CVF Conference on Computer Vision and
  Pattern Recognition}, 2022.

\bibitem[\protect\citeauthoryear{Li \bgroup \em et al.\egroup
  }{2022b}]{li2022toward}
Hao Li, Jinfa Huang, Peng Jin, Guoli Song, Qi~Wu, and Jie Chen.
\newblock Toward 3d spatial reasoning for human-like text-based visual question
  answering.
\newblock {\em arXiv preprint arXiv:2209.10326}, 2022.

\bibitem[\protect\citeauthoryear{Li \bgroup \em et al.\egroup
  }{2022c}]{li2022joint}
Hao Li, Xu~Li, Belhal Karimi, Jie Chen, and Mingming Sun.
\newblock Joint learning of object graph and relation graph for visual question
  answering.
\newblock In {\em 2022 IEEE International Conference on Multimedia and Expo
  (ICME)}, pages 01--06. IEEE, 2022.

\bibitem[\protect\citeauthoryear{Li \bgroup \em et al.\egroup
  }{2022d}]{li2022fine}
Juncheng Li, Xin He, Longhui Wei, Long Qian, Linchao Zhu, Lingxi Xie, Yueting
  Zhuang, Qi~Tian, and Siliang Tang.
\newblock Fine-grained semantically aligned vision-language pre-training.
\newblock {\em arXiv preprint arXiv:2208.02515}, 2022.

\bibitem[\protect\citeauthoryear{Li \bgroup \em et al.\egroup
  }{2022e}]{li2022dynamic}
Kehan Li, Zhennan Wang, Zesen Cheng, Runyi Yu, Yian Zhao, Guoli Song, Li~Yuan,
  and Jie Chen.
\newblock Dynamic clustering network for unsupervised semantic segmentation.
\newblock {\em arXiv preprint arXiv:2210.05944}, 2022.

\bibitem[\protect\citeauthoryear{Li \bgroup \em et al.\egroup
  }{2022f}]{li2022invariant}
Yicong Li, Xiang Wang, Junbin Xiao, Wei Ji, and Tat-Seng Chua.
\newblock Invariant grounding for video question answering.
\newblock In {\em Proceedings of the IEEE/CVF Conference on Computer Vision and
  Pattern Recognition}, pages 2928--2937, 2022.

\bibitem[\protect\citeauthoryear{Li \bgroup \em et al.\egroup
  }{2023}]{li2023multi}
Kehan Li, Yian Zhao, Zhennan Wang, Zesen Cheng, Peng Jin, Xiangyang Ji,
  Li~Yuan, Chang Liu, and Jie Chen.
\newblock Multi-granularity interaction simulation for unsupervised interactive
  segmentation.
\newblock {\em arXiv preprint arXiv:2303.13399}, 2023.

\bibitem[\protect\citeauthoryear{Luo \bgroup \em et al.\egroup
  }{2022}]{luo2022clip4clip}
Huaishao Luo, Lei Ji, Ming Zhong, Yang Chen, Wen Lei, Nan Duan, and Tianrui Li.
\newblock Clip4clip: An empirical study of clip for end to end video clip
  retrieval and captioning.
\newblock {\em Neurocomputing}, 508:293--304, 2022.

\bibitem[\protect\citeauthoryear{Making}{2009}]{makingsynthesis}
Making.
\newblock Synthesis lectures on artificial intelligence and machine learning.
\newblock 2009.

\bibitem[\protect\citeauthoryear{Marichal and
  Mathonet}{2011}]{marichal2011weighted}
Jean-Luc Marichal and Pierre Mathonet.
\newblock Weighted banzhaf power and interaction indexes through weighted
  approximations of games.
\newblock {\em European journal of operational research}, 2011.

\bibitem[\protect\citeauthoryear{Peng \bgroup \em et al.\egroup
  }{2022}]{peng2022multilevel}
Min Peng, Chongyang Wang, Yuan Gao, Yu~Shi, and Xiang-Dong Zhou.
\newblock Multilevel hierarchical network with multiscale sampling for video
  question answering.
\newblock {\em arXiv preprint arXiv:2205.04061}, 2022.

\bibitem[\protect\citeauthoryear{Piergiovanni \bgroup \em et al.\egroup
  }{2022}]{piergiovanni2022video}
AJ~Piergiovanni, Kairo Morton, Weicheng Kuo, Michael~S Ryoo, and Anelia
  Angelova.
\newblock Video question answering with iterative video-text co-tokenization.
\newblock In {\em European Conference on Computer Vision}, pages 76--94.
  Springer, 2022.

\bibitem[\protect\citeauthoryear{Qian \bgroup \em et al.\egroup
  }{2022}]{qian2022locate}
Tianwen Qian, Ran Cui, Jingjing Chen, Pai Peng, Xiaowei Guo, and Yu-Gang Jiang.
\newblock Locate before answering: Answer guided question localization for
  video question answering.
\newblock {\em arXiv:2210.02081}, 2022.

\bibitem[\protect\citeauthoryear{Radford \bgroup \em et al.\egroup
  }{2018}]{radford2018improving}
Alec Radford, Karthik Narasimhan, Tim Salimans, Ilya Sutskever, et~al.
\newblock Improving language understanding by generative pre-training.
\newblock 2018.

\bibitem[\protect\citeauthoryear{Radford \bgroup \em et al.\egroup
  }{2021}]{radford2021learning}
Alec Radford, Jong~Wook Kim, Chris Hallacy, Aditya Ramesh, Gabriel Goh,
  Sandhini Agarwal, Girish Sastry, Amanda Askell, Pamela Mishkin, Jack Clark,
  et~al.
\newblock Learning transferable visual models from natural language
  supervision.
\newblock In {\em International Conference on Machine Learning}, pages
  8748--8763. PMLR, 2021.

\bibitem[\protect\citeauthoryear{Raffel \bgroup \em et al.\egroup
  }{2020}]{raffel2020exploring}
Colin Raffel, Noam Shazeer, Adam Roberts, Katherine Lee, Sharan Narang, Michael
  Matena, Yanqi Zhou, Wei Li, Peter~J Liu, et~al.
\newblock Exploring the limits of transfer learning with a unified text-to-text
  transformer.
\newblock {\em J. Mach. Learn. Res.}, 21(140):1--67, 2020.

\bibitem[\protect\citeauthoryear{Rodriguez and
  Laio}{2014}]{rodriguez2014clustering}
Alex Rodriguez and Alessandro Laio.
\newblock Clustering by fast search and find of density peaks.
\newblock {\em science}, 344(6191):1492--1496, 2014.

\bibitem[\protect\citeauthoryear{Seo \bgroup \em et al.\egroup
  }{2021}]{seo2021look}
Paul~Hongsuck Seo, Arsha Nagrani, and Cordelia Schmid.
\newblock Look before you speak: Visually contextualized utterances.
\newblock In {\em Proceedings of the IEEE/CVF Conference on Computer Vision and
  Pattern Recognition}, pages 16877--16887, 2021.

\bibitem[\protect\citeauthoryear{Sun \bgroup \em et al.\egroup
  }{2020}]{sun2020random}
Jianyuan Sun, Hui Yu, Guoqiang Zhong, Junyu Dong, Shu Zhang, and Hongchuan Yu.
\newblock Random shapley forests: cooperative game-based random forests with
  consistency.
\newblock {\em IEEE transactions on cybernetics}, 2020.

\bibitem[\protect\citeauthoryear{Sun \bgroup \em et al.\egroup
  }{2021}]{sun2021video}
Guanglu Sun, Lili Liang, Tianlin Li, Bo~Yu, Meng Wu, and Bolun Zhang.
\newblock Video question answering: a survey of models and datasets.
\newblock {\em Mobile Networks and Applications}, 26(5):1904--1937, 2021.

\bibitem[\protect\citeauthoryear{Tang \bgroup \em et al.\egroup
  }{2021}]{tang2021discovering}
Zhenggang Tang, Chao Yu, Boyuan Chen, Huazhe Xu, Xiaolong Wang, Fei Fang, Simon
  Du, Yu~Wang, and Yi~Wu.
\newblock Discovering diverse multi-agent strategic behavior via reward
  randomization.
\newblock {\em arXiv preprint arXiv:2103.04564}, 2021.

\bibitem[\protect\citeauthoryear{Wu \bgroup \em et al.\egroup
  }{2017}]{wu2017visual}
Qi~Wu, Damien Teney, Peng Wang, Chunhua Shen, Anthony Dick, and Anton Van
  Den~Hengel.
\newblock Visual question answering: A survey of methods and datasets.
\newblock {\em Computer Vision and Image Understanding}, 163:21--40, 2017.

\bibitem[\protect\citeauthoryear{Xiao \bgroup \em et al.\egroup
  }{2021}]{xiao2021next}
Junbin Xiao, Xindi Shang, Angela Yao, and Tat-Seng Chua.
\newblock Next-qa: Next phase of question-answering to explaining temporal
  actions.
\newblock In {\em Proceedings of the IEEE/CVF Conference on Computer Vision and
  Pattern Recognition}, pages 9777--9786, 2021.

\bibitem[\protect\citeauthoryear{Xie \bgroup \em et al.\egroup
  }{2018}]{xie2018rethinking}
Saining Xie, Chen Sun, Jonathan Huang, Zhuowen Tu, and Kevin Murphy.
\newblock Rethinking spatiotemporal feature learning: Speed-accuracy trade-offs
  in video classification.
\newblock In {\em Proceedings of the European conference on computer vision
  (ECCV)}, pages 305--321, 2018.

\bibitem[\protect\citeauthoryear{Xu \bgroup \em et al.\egroup
  }{2017}]{xu2017video}
Dejing Xu, Zhou Zhao, Jun Xiao, Fei Wu, Hanwang Zhang, Xiangnan He, and Yueting
  Zhuang.
\newblock Video question answering via gradually refined attention over
  appearance and motion.
\newblock In {\em Proceedings of the 25th ACM Multimedia}, pages 1645--1653,
  2017.

\bibitem[\protect\citeauthoryear{Yang \bgroup \em et al.\egroup
  }{2022a}]{yang2022learning}
Antoine Yang, Antoine Miech, Josef Sivic, Ivan Laptev, and Cordelia Schmid.
\newblock Learning to answer visual questions from web videos.
\newblock {\em arXiv preprint arXiv:2205.05019}, 2022.

\bibitem[\protect\citeauthoryear{Yang \bgroup \em et al.\egroup
  }{2022b}]{yang2022zero}
Antoine Yang, Antoine Miech, Josef Sivic, Ivan Laptev, and Cordelia Schmid.
\newblock Zero-shot video question answering via frozen bidirectional language
  models.
\newblock {\em arXiv preprint arXiv:2206.08155}, 2022.

\bibitem[\protect\citeauthoryear{Ye \bgroup \em et al.\egroup
  }{2023}]{ye2023fits}
Qichen Ye, Bowen Cao, Nuo Chen, Weiyuan Xu, and Yuexian Zou.
\newblock Fits: Fine-grained two-stage training for knowledge-aware question
  answering.
\newblock {\em arXiv preprint arXiv:2302.11799}, 2023.

\bibitem[\protect\citeauthoryear{Yu \bgroup \em et al.\egroup
  }{2019}]{yu2019activitynet}
Zhou Yu, Dejing Xu, Jun Yu, Ting Yu, Zhou Zhao, Yueting Zhuang, and Dacheng
  Tao.
\newblock Activitynet-qa: A dataset for understanding complex web videos via
  question answering.
\newblock In {\em Proceedings of the AAAI Conference on Artificial
  Intelligence}, volume~33, pages 9127--9134, 2019.

\bibitem[\protect\citeauthoryear{Zhong \bgroup \em et al.\egroup
  }{2022}]{zhong2022video}
Yaoyao Zhong, Wei Ji, Junbin Xiao, Yicong Li, Weihong Deng, and Tat-Seng Chua.
\newblock Video question answering: Datasets, algorithms and challenges.
\newblock {\em arXiv preprint arXiv:2203.01225}, 2022.

\end{thebibliography}

\end{document}